\documentclass{article} 
\usepackage{iclr2023_conference_tinypaper,times}
\iclrfinalcopy

\usepackage{amsmath,amsfonts,bm}

\def\eqref#1{equation~\ref{#1}}

\def\1{\bm{1}}

\DeclareMathAlphabet{\mathsfit}{\encodingdefault}{\sfdefault}{m}{sl}
\SetMathAlphabet{\mathsfit}{bold}{\encodingdefault}{\sfdefault}{bx}{n}

\usepackage{hyperref}
\usepackage{url}
\usepackage{dirtytalk}
\usepackage{amssymb} 
\usepackage{mathrsfs} 
\usepackage{graphicx} 
\usepackage{wrapfig} 
\usepackage{subcaption} 
\usepackage{float}

\usepackage{amsthm}
\newtheorem{definition}{Definition}[section]
\newtheorem{example}{Example}[section]

\title{
Partial Rankings of Optimizers}

\author{Julian Rodemann$^*$, Hannah Blocher$^*$\\
Department of Statistics, 
Ludwig-Maximilians-Universität München, Munich, Germany \\
\texttt{\{julian.rodemann,hannah.blocher\}@stat.uni-muenchen.de} \\
}

\newcommand{\Pcal}{\mathcal{P}}

\newcommand{\Sscr}{\mathscr{S}}

\begin{document}

\def\thefootnote{*}\footnotetext{Authors contributed equally to this work. Total order was enforced by fair coin flip.}\def\thefootnote{\arabic{footnote}}

\maketitle
\begin{abstract}
We introduce a framework for benchmarking optimizers according to multiple criteria over a collection of test functions. Based on a recently introduced union-free generic depth function for partial orders/rankings, it fully exploits the ordinal information and allows for incomparability. Our method describes the distribution of all partial orders/rankings, avoiding the notorious shortcomings of aggregation. This permits to identify test functions that produce central or outlying rankings of optimizers and to assess the quality of benchmarking suites.\footnote{\textbf{Code:} \url{https://github.com/hannahblo/Posets_Optimizers}}

\end{abstract}

\section{Introduction}

Despite its importance for machine learning research, there is no broad agreement on how to compare optimization algorithms on benchmark suites with regard to multiple criteria, see \cite{hansen2022anytime} for instance. This is particularly relevant for multi-objective optimization, which has diverse applications ranging from 
reinforcement learning \citep{basaklar2023pdmorl,zhu2023scaling} to representation learning \citep{gu2023min}, neural architecture search \citep{lu2019nsga} and large language models \citep{anonymous2023beyond}. But such comparisons also arise when single-objective optimizers are evaluated with respect to several metrics, see \cite{pmlr-v119-sivaprasad20a, MLSYS2020_411e39b1, dahl2023benchmarking}. A popular example is the duality of fixed-budget (performance) and fixed-target (speed) evaluation of deep learning optimizers, see e.g. \cite{dewancker2016strategy} or results in section~\ref{sec:results}.

We propose a novel framework for comparing optimizers with respect to multiple criteria over a benchmarking suite of test functions. It is motivated by two observations from benchmarking practice. Firstly, in many cases, ranking optimizers is the overall aim of benchmarking, which then renders the metric information a mere means to an end. Secondly, multiple criteria give rise to incomparability of optimizers (\say{better with respect to one metric, worse with respect to another one}) -- a fact that classical aggregation methods aiming at \textit{total} orders fail to represent \citep{dewancker2016strategy}. In general, it has been proven that it is impossible to aggregate a set of total orders to a single total order while assuming natural conditions on the aggregation, see appendix~\ref{app: related work}. In contrast, 
our framework is based on \textit{partially} ordering optimizers according to their performance on a single test function, thus deliberately allowing for incomparability. When considering a benchmarking suite of test functions, we obtain a set of such partially ordered sets (posets) describing the performance of optimizers. To evaluate these posets, we use the concept of depth functions which provide a notion of centrality and outlyingness, see~\cite{zuo2000general} and the adaptation to poset-valued data developed by~\cite{pmlr-v215-blocher23a} which is called \textit{union-free generic (ufg) depth}. This gives us a description of the distribution and not just an aggregation. Thus, by applying the ufg depth to the posets given by a benchmark suite, we can identify those test functions that give a central/well-supported performance ordering of the optimizers and those test functions that return outliers, see appendices~\ref{results-bbob} and~\ref{sec: results-multi-objective-EA}. 
This paves the way for analyzing the diversity of problems covered by benchmarking suites. 

\section{Method}\label{sec:method}

Depth functions describe an empirical distribution by indicating how central/outlying each point is relative to an entire data cloud or underlying distribution. Our framework relies on the union-free generic (ufg) depth presented in~\cite{pmlr-v215-blocher23a}. This is an adaptation of the simplicial depth function, which denotes the probability that a point $x \in \mathbb{R}^d$ (with $d \in \mathbb{N}$) lies in a randomly drawn $d+1$ simplex (,i.e., in the output of the convex hull/closure operator, see appendix~\ref{app: prem_def}). For adaptation to posets, let $\Pcal$ be the set of all possible posets on a finite set $M$. The ufg depth is based on the closure operator $\gamma\colon 2^{\Pcal} \to 2^{\Pcal}, P \mapsto \left\{p \in \Pcal \mid \cap_{\tilde{p}\in P}\: \tilde{p} \subseteq p \subseteq \cup_{\tilde{p} \in P}\:\tilde{p} \right\}.$ Analogous to the simplicial depth, where only the edges of the $d+1$ simplices are used, the ufg depth considers only those subsets of $2^{\Pcal}$ that are non-trivial, minimal, and not decomposable without loss of information based on the closure operator $\gamma$. More specifically, set $\Sscr = \left\{P \subseteq\Pcal \mid \text{ condition } (C1) \text{ and } (C2) \text{ hold} \right\}$ with (C1): $P \subsetneq \gamma(P),$ and (C2): there exists no family $(A_i)_{i \in \{1, \ldots, \ell\}}$ such that for all $i \in \{1, \ldots, \ell\}$ $A_i\subsetneq P$ and $\cup_{i \in \{1, \ldots, \ell\}}\gamma(A_i) = \gamma(P)$. Thus, the ufg depth of a poset $p\in \Pcal$ is the proportion of sets $S \in \Sscr$ that contain (i.e. $p \in \gamma(S)$) the poset of interest. More details 
can be found in appendix~\ref{app:def ufg depth}.
Consider now the task of comparing $d$ optimizers with respect to $c$ criteria for each of $n$ test functions. 
Based on a fixed test function, we say that optimizer $i$ is better than optimizer $j$ iff optimizer $i$ is better for at least one criterion and not worse for all others. Thus, if there are two criteria that contradict each other (one criterion states that optimizer $i$ performs better than optimizer $j$ and the other criterion states the opposite), then we say that these two optimizers are incomparable.\footnote{Equality of all criteria values does not occur in the following illustration and is discussed in appendix~\ref{results-bbob}.} Hence, when we consider $n$ test functions, we obtain $n$ posets describing the performance of $d$ optimizers based on $c$ criteria. 
Now, we apply the ufg depth on these posets. Then, the poset with the highest ufg depth value contains the structure that is most supported by the performance order given by all the test functions, while the poset with the lowest ufg depth gives a structure that can be seen as an outlier. Another interpretation is that a test function that produces the poset with a high (low) ufg depth value is a typical (an atypical) problem compared to the other test functions.

\section{Results}\label{sec:results}

\begin{wrapfigure}{r}{0.33\textwidth}
\vspace{-0.7cm}
  \begin{center}
    \includegraphics[width=0.33\textwidth]{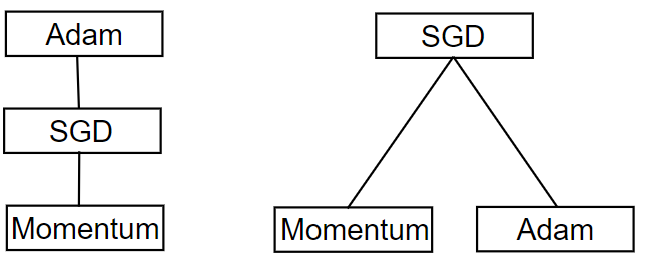}
  \end{center}
  \caption{Orderings of optimizers corresponding to highest ($0.65$, left) and lowest ($0.29$, right) ufg depth.}
  \label{fig:max_min_depth_poset:main}
  \vspace{-0.3cm}
\end{wrapfigure}

We illustrate our framework on \texttt{DeepOBS}, a benchmark suite for deep learning optimizers \citep{schneider2019deepobs}. We closely mimic the setup in \citet[section 4]{schneider2019deepobs} and compare vanilla stochastic gradient descent (SGD), adam, and momentum as baselines on 8 test functions, which arise from training various models on different data sets, e.g., a Long Short-Term Memory network (LSTM) \citep{hochreiter1997long} for character-level language modeling on Leo Tolstoi’s \textit{War and Peace}, see \citet[Appendix A]{schneider2019deepobs} for details. We benchmark SGD, adam, and momentum with respect to performance (minimal test loss achieved in a fixed time budget) and speed (time required to achieve a given test loss).\footnote{Further results for 11 optimizers on the Black-box Optimization Benchmarking (BBOB) suite \citep{hansen2010comparing} comprising 24 test functions as well as for 7 multi-objective evolutionary algorithms \citep{wu2023dynamic} benchmarked on 13 test functions can be found in appendix \ref{results-bbob} and \ref{sec: results-multi-objective-EA}, respectively. 
}
We obtain 8 posets describing the order of the optimizers. We observe 5 unique and 3 duplicated posets. From the 8 observed posets, figure~\ref{fig:max_min_depth_poset:main} shows the poset corresponding to maximal and minimal ufg depth value. Here, an ascending chain of edges between any two optimizers means that the optimizer below is outperformed by the one above. The poset on the left has the highest ufg depth value with 0.65 and therefore the structure which is most supported by the observed benchmarking results. This poset is the duplicated one. In contrast, the poset on the right with the lowest ufg depth value of 0.29 can be seen as outlying. This means that the underlying problem (LSTM on \textit{War and Peace}) produces an order structure that is atypical compared to the other 7 orders given by the test functions.
Indeed, it appears surprising for vanilla SGD to outperform adam and momentum, the latter two being enhancements of SGD.
We reckon the reason could be that the other 7 test functions in \texttt{DeepOBS} are all based on more modern network architectures than LSTM. 
Besides delivering benchmarking results, our framework also informs the benchmarking suite \textit{designer} about the distribution of orderings produced by her suite. Depending on its overall aim, the designer might use this information to remove or add test functions, giving rise to a more targeted curation of benchmarking suites.
For instance, the detection of LSTM on \textit{War and Peace} as an outlier in \texttt{DeepOBS} could lead to the removal of the latter in case the designer wants the suite to test optimizers on modern network architectures, and vice versa. 
See appendix \ref{app:benchmark-suite-design} for more details on how our framework can inform benchmarking suite designers.

\section{URM Statement}

The authors acknowledge that at least one key author of this work meets the URM criteria of ICLR 2024 Tiny Papers Track. 

 \section{Acknowledgements}
We sincerely thank Olaf Mersmann for sharing the BBOB results data with us. Another heartfelt thanks goes to Sebastian Fischer and Christoph Jansen for comments on earlier versions of this manuscript. We are also grateful to the three anonymous reviewers for their review of the first version of this manuscript. JR acknowledges support by the Bavarian Academy of Sciences (BAS) through the Bavarian Institute for Digital Transformation (bidt) and by the Federal Statistical Office of Germany within the co-operation project "Machine Learning in Official Statistics". HB sincerely thanks Evangelisches Studienwerk Villigst e.V. for funding and supporting her doctoral studies.

Both authors acknowledge support by the LMU mentoring program.


\bibliography{bib}

\begin{thebibliography}{36}
\providecommand{\natexlab}[1]{#1}
\providecommand{\url}[1]{\texttt{#1}}
\expandafter\ifx\csname urlstyle\endcsname\relax
  \providecommand{\doi}[1]{doi: #1}\else
  \providecommand{\doi}{doi: \begingroup \urlstyle{rm}\Url}\fi

\bibitem[Arrow(1950)]{arrow1950difficulty}
Kenneth~J Arrow.
\newblock A difficulty in the concept of social welfare.
\newblock \emph{Journal of political economy}, 58\penalty0 (4):\penalty0 328--346, 1950.

\bibitem[Auger \& Hansen(2005)Auger and Hansen]{auger05}
Anne Auger and Nikolaus Hansen.
\newblock Performance evaluation of an advanced local search evolutionary algorithm.
\newblock In \emph{IEEE Congress on Evolutionary Computation}. IEEE, 2005.

\bibitem[Bacharach(1975)]{bacharach1975group}
Michael Bacharach.
\newblock Group decisions in the face of differences of opinion.
\newblock \emph{Management Science}, 22\penalty0 (2):\penalty0 182--191, 1975.

\bibitem[Basaklar et~al.(2023)Basaklar, Gumussoy, and Ogras]{basaklar2023pdmorl}
Toygun Basaklar, Suat Gumussoy, and Umit Ogras.
\newblock {PD}-{MORL}: Preference-driven multi-objective reinforcement learning algorithm.
\newblock In \emph{The Eleventh International Conference on Learning Representations (ICLR)}, 2023.

\bibitem[Blocher et~al.(2023)Blocher, Schollmeyer, Jansen, and Nalenz]{pmlr-v215-blocher23a}
Hannah Blocher, Georg Schollmeyer, Christoph Jansen, and Malte Nalenz.
\newblock Depth functions for partial orders with a descriptive analysis of machine learning algorithms.
\newblock In \emph{Proceedings of the Thirteenth International Symposium on Imprecise Probability: Theories and Applications (ISIPTA)}, volume 215, pp.\  59--71. PMLR, 2023.

\bibitem[Borda(1781)]{borda1781memoire}
Jean Charles~de Borda.
\newblock M{\'e}moire sur les {\'e}lections au scrutin.
\newblock \emph{Histoire de l’Acad{\'e}mie Royale des Sciences}, 12, 1781.

\bibitem[Condorcet(1785)]{marquis85}
Marquis~de Condorcet.
\newblock Essai sur l’application de l’analyse a la probabilite des decisions rendues a la pluralite des voix, 1785.
\newblock Paris.

\bibitem[Dahl et~al.(2023)Dahl, Schneider, Nado, Agarwal, Sastry, Hennig, Medapati, Eschenhagen, Kasimbeg, Suo, et~al.]{dahl2023benchmarking}
George~E Dahl, Frank Schneider, Zachary Nado, Naman Agarwal, Chandramouli~Shama Sastry, Philipp Hennig, Sourabh Medapati, Runa Eschenhagen, Priya Kasimbeg, Daniel Suo, et~al.
\newblock Benchmarking neural network training algorithms.
\newblock \emph{arXiv preprint arXiv:2306.07179}, 2023.

\bibitem[Dewancker et~al.(2016)Dewancker, McCourt, Clark, Hayes, Johnson, and Ke]{dewancker2016strategy}
Ian Dewancker, Michael McCourt, Scott Clark, Patrick Hayes, Alexandra Johnson, and George Ke.
\newblock A strategy for ranking optimization methods using multiple criteria.
\newblock In \emph{Workshop on Automatic Machine Learning}, pp.\  11--20. PMLR, 2016.

\bibitem[Dwork et~al.(2001)Dwork, Kumar, Naor, and Sivakumar]{dwork2001rank}
Cynthia Dwork, Ravi Kumar, Moni Naor, and Dandapani Sivakumar.
\newblock Rank aggregation methods for the web.
\newblock In \emph{Proceedings of the 10th international conference on World Wide Web}, pp.\  613--622, 2001.

\bibitem[Eckhoff(1993)]{eckhoff93}
Jürgen Eckhoff.
\newblock Chapter 2.1 - {H}elly, {R}adon, and {C}arathéodory type theorems.
\newblock In \emph{Handbook of Convex Geometry}, pp.\  389--448. North-Holland, Amsterdam, 1993.

\bibitem[French \& Insua(2010)French and Insua]{simon10}
Simon French and David~Rios Insua.
\newblock \emph{Statistical Decision Theory: Kendall's Library of Statistics 9}.
\newblock Wiley, 2010.

\bibitem[Gijbels \& Nagy(2017)Gijbels and Nagy]{gijbels17}
Ir{\`e}ne Gijbels and Stanislav Nagy.
\newblock On a general definition of depth for functional data.
\newblock \emph{Statistical Science}, 32\penalty0 (4):\penalty0 630--639, 2017.

\bibitem[Gu et~al.(2023)Gu, Lu, Ram, and Weng]{gu2023min}
Alex Gu, Songtao Lu, Parikshit Ram, and Lily Weng.
\newblock Min-max multi-objective bilevel optimization with applications in robust machine learning.
\newblock In \emph{International Conference on Learning Representations (ICLR)}, 2023.

\bibitem[Hansen et~al.(2010)Hansen, Auger, Ros, Finck, and Po{\v{s}}{\'\i}k]{hansen2010comparing}
Nikolaus Hansen, Anne Auger, Raymond Ros, Steffen Finck, and Petr Po{\v{s}}{\'\i}k.
\newblock Comparing results of 31 algorithms from the black-box optimization benchmarking bbob-2009.
\newblock In \emph{Proceedings of the 12th annual conference companion on Genetic and evolutionary computation}, pp.\  1689--1696, 2010.

\bibitem[Hansen et~al.(2021)Hansen, Auger, Ros, Mersmann, Tu{\v s}ar, and Brockhoff]{hansen2021coco}
Nikolaus Hansen, Anne Auger, Raymond Ros, Olaf Mersmann, Tea Tu{\v s}ar, and Dimo Brockhoff.
\newblock {COCO}: A platform for comparing continuous optimizers in a black-box setting.
\newblock \emph{Optimization Methods and Software}, 36:\penalty0 114--144, 2021.
\newblock URL \url{http://numbbo.github.io/coco/}.
\newblock (accessed: 08.12.2023).

\bibitem[Hansen et~al.(2022)Hansen, Auger, Brockhoff, and Tu{\v{s}}ar]{hansen2022anytime}
Nikolaus Hansen, Anne Auger, Dimo Brockhoff, and Tea Tu{\v{s}}ar.
\newblock Anytime performance assessment in blackbox optimization benchmarking.
\newblock \emph{IEEE Transactions on Evolutionary Computation}, 26\penalty0 (6):\penalty0 1293--1305, 2022.

\bibitem[Hochreiter \& Schmidhuber(1997)Hochreiter and Schmidhuber]{hochreiter1997long}
Sepp Hochreiter and J{\"u}rgen Schmidhuber.
\newblock Long short-term memory.
\newblock \emph{Neural computation}, 9\penalty0 (8):\penalty0 1735--1780, 1997.

\bibitem[Jansen et~al.(2018{\natexlab{a}})Jansen, Schollmeyer, and Augustin]{jansen18}
Christoph Jansen, Georg Schollmeyer, and Thomas Augustin.
\newblock A probabilistic evaluation framework for preference aggregation reflecting group homogeneity.
\newblock \emph{Mathematical Social Sciences}, \penalty0 (96):\penalty0 49--62, 2018{\natexlab{a}}.

\bibitem[Jansen et~al.(2018{\natexlab{b}})Jansen, Schollmeyer, and Augustin]{jsa2018}
Christoph Jansen, Georg Schollmeyer, and Thomas Augustin.
\newblock Concepts for decision making under severe uncertainty with partial ordinal and partial cardinal preferences.
\newblock \emph{International Journal of Approximate Reasoning}, 98:\penalty0 112--131, 2018{\natexlab{b}}.

\bibitem[Jansen et~al.(2023{\natexlab{a}})Jansen, Nalenz, Schollmeyer, and Augustin]{jansen2023statistical}
Christoph Jansen, Malte Nalenz, Georg Schollmeyer, and Thomas Augustin.
\newblock Statistical comparisons of classifiers by generalized stochastic dominance.
\newblock \emph{Journal of Machine Learning Research}, 24\penalty0 (231):\penalty0 1--37, 2023{\natexlab{a}}.

\bibitem[Jansen et~al.(2023{\natexlab{b}})Jansen, Schollmeyer, Blocher, Rodemann, and Augustin]{uaiall}
Christoph Jansen, Georg Schollmeyer, Hannah Blocher, Julian Rodemann, and Thomas Augustin.
\newblock Robust statistical comparison of random variables with locally varying scale of measurement.
\newblock In Robin~J. Evans and Ilya Shpitser (eds.), \emph{Proceedings of the Thirty-Ninth Conference on Uncertainty in Artificial Intelligence}, volume 216 of \emph{Proceedings of Machine Learning Research}, pp.\  941--952. PMLR, 31 Jul--04 Aug 2023{\natexlab{b}}.

\bibitem[Kemeny \& Snell(1962)Kemeny and Snell]{kemeny62}
John~G. Kemeny and J~Laurie Snell.
\newblock Preference ranking: An axiomatic approach. mathematical.
\newblock \emph{Mathematical Models in Social Science.}, pp.\  9--23, 1962.

\bibitem[Kleitman \& Rothschild(1970)Kleitman and Rothschild]{kleitman70}
Daniel Kleitman and Bruce Rothschild.
\newblock The number of finite topologies.
\newblock \emph{Proceedings of the American Mathematical Society}, 25\penalty0 (2):\penalty0 276, 1970.

\bibitem[Liu(1990)]{liu90}
Regina Liu.
\newblock On a notion of data depth based on random simplices.
\newblock \emph{The Annals of Statistics}, 18:\penalty0 405--414, 1990.

\bibitem[Lu et~al.(2019)Lu, Whalen, Boddeti, Dhebar, Deb, Goodman, and Banzhaf]{lu2019nsga}
Zhichao Lu, Ian Whalen, Vishnu Boddeti, Yashesh Dhebar, Kalyanmoy Deb, Erik Goodman, and Wolfgang Banzhaf.
\newblock Nsga-net: neural architecture search using multi-objective genetic algorithm.
\newblock In \emph{Proceedings of the genetic and evolutionary computation conference}, pp.\  419--427, 2019.

\bibitem[Mattson et~al.(2020)Mattson, Cheng, Diamos, Coleman, Micikevicius, Patterson, Tang, Wei, Bailis, Bittorf, Brooks, Chen, Dutta, Gupta, Hazelwood, Hock, Huang, Kang, Kanter, Kumar, Liao, Narayanan, Oguntebi, Pekhimenko, Pentecost, Janapa~Reddi, Robie, St~John, Wu, Xu, Young, and Zaharia]{MLSYS2020_411e39b1}
Peter Mattson, Christine Cheng, Gregory Diamos, Cody Coleman, Paulius Micikevicius, David Patterson, Hanlin Tang, Gu-Yeon Wei, Peter Bailis, Victor Bittorf, David Brooks, Dehao Chen, Debo Dutta, Udit Gupta, Kim Hazelwood, Andy Hock, Xinyuan Huang, Daniel Kang, David Kanter, Naveen Kumar, Jeffery Liao, Deepak Narayanan, Tayo Oguntebi, Gennady Pekhimenko, Lillian Pentecost, Vijay Janapa~Reddi, Taylor Robie, Tom St~John, Carole-Jean Wu, Lingjie Xu, Cliff Young, and Matei Zaharia.
\newblock {MLP}erf training benchmark.
\newblock In \emph{Proceedings of Machine Learning and Systems}, volume~2, pp.\  336--349, 2020.

\bibitem[Mersmann et~al.(2010)Mersmann, Trautmann, Naujoks, and Weihs]{mersmann2010benchmarking}
Olaf Mersmann, Heike Trautmann, Boris Naujoks, and Claus Weihs.
\newblock Benchmarking evolutionary multiobjective optimization algorithms.
\newblock In \emph{IEEE Congress on Evolutionary Computation}, pp.\  1--8. IEEE, 2010.

\bibitem[Mersmann et~al.(2015)Mersmann, Preuss, Trautmann, Bischl, and Weihs]{mersmann15}
Olaf Mersmann, Mike Preuss, Heike Trautmann, Bernd Bischl, and Claus Weihs.
\newblock Analyzing the bbob results by means of benchmarking concepts.
\newblock \emph{Evolutionary Computation}, 23\penalty0 (1):\penalty0 161--185, 2015.

\bibitem[Schneider et~al.(2019)Schneider, Balles, and Hennig]{schneider2019deepobs}
Frank Schneider, Lukas Balles, and Philipp Hennig.
\newblock Deepobs: A deep learning optimizer benchmark suite.
\newblock In \emph{International Conference on Learning Representations (ICLR)}, 2019.
\newblock URL \url{https://deepobs.github.io/}.
\newblock (accessed: 08.12.2023).

\bibitem[Sivaprasad et~al.(2020)Sivaprasad, Mai, Vogels, Jaggi, and Fleuret]{pmlr-v119-sivaprasad20a}
Prabhu~Teja Sivaprasad, Florian Mai, Thijs Vogels, Martin Jaggi, and Fran{\c{c}}ois Fleuret.
\newblock Optimizer benchmarking needs to account for hyperparameter tuning.
\newblock In \emph{Proceedings of the 37th International Conference on Machine Learning (ICML)}, volume 119, pp.\  9036--9045. PMLR, 2020.

\bibitem[Wu et~al.(2023)Wu, Wang, Chen, and Wang]{wu2023dynamic}
Fei Wu, Wanliang Wang, Jiacheng Chen, and Zheng Wang.
\newblock A dynamic multi-objective optimization method based on classification strategies.
\newblock \emph{Scientific Reports}, 13\penalty0 (1):\penalty0 15221, 2023.

\bibitem[Yannakakis(1982)]{yannakakis82}
Mihalis Yannakakis.
\newblock The complexity of the partial order dimension problem.
\newblock \emph{SIAM Journal on Algebraic Discrete Methods}, 3\penalty0 (3):\penalty0 351--358, 1982.

\bibitem[Zhou et~al.(2023)Zhou, Liu, Yang, Shao, Liu, Yue, Ouyang, and Qiao]{anonymous2023beyond}
Zhanhui Zhou, Jie Liu, Chao Yang, Jing Shao, Yu~Liu, Xiangyu Yue, Wanli Ouyang, and Yu~Qiao.
\newblock Beyond one-preference-for-all: Multi-objective direct preference optimization.
\newblock \emph{arXiv preprint arXiv:2310.03708}, 2023.

\bibitem[Zhu et~al.(2023)Zhu, Dang, and Grover]{zhu2023scaling}
Baiting Zhu, Meihua Dang, and Aditya Grover.
\newblock Scaling pareto-efficient decision making via offline multi-objective {RL}.
\newblock In \emph{The Eleventh International Conference on Learning Representations (ICLR)}, 2023.

\bibitem[Zuo \& Serfling(2000)Zuo and Serfling]{zuo2000general}
Yijun Zuo and Robert Serfling.
\newblock General notions of statistical depth function.
\newblock \emph{Annals of statistics}, pp.\  461--482, 2000.

\end{thebibliography}
\bibliographystyle{iclr2023_conference_tinypaper}

\newpage
\appendix
\section{Preliminary Definitions}\label{app: prem_def}
In this section, we state all general definitions necessary to this paper.

\textit{Partial orders (posets)} are based on a fixed set $M$, which orders the elements of $M$. A partial order $p$ is therefore a subset of $M\times M$ which is reflexive (for all $y \in M$ we have $(y,y) \in p$), antisymmetric (if $(y_1, y_2) \in p$ with $y_1 \neq y_2$ then $(y_2, y_1) \not\in p$) and transitive (if $(y_1, y_2),(y_2,y_3) \in p$ then $(y_1, y_3) \in p$). A poset that is strongly connected (for all $y_1, y_2 \in M$ either $(y_1, y_2)\in p$ or $(y_2, y_1) \in p$ holds) is called a \textit{total/linear order}. Note that a subset of $M\times M$ that is only reflexive and transitive is called a \textit{preorder}. 

A \textit{closure operator} on a set $\Omega$ is a function $\gamma_{\Omega}: 2^{\Omega} \to 2^{\Omega}$ which is extensive (for $A \subseteq \Omega$ we have $A \subseteq \gamma_{\Omega}(A)$), idempotent (for $A \subseteq \Omega$ we have $\gamma_{\Omega}(A) = \gamma_{\Omega}(\gamma_{\Omega}(A))$), and increasing (for $A, B \subseteq \Omega$ with $A \subseteq B$ we have $\gamma_{\Omega}(A) \subseteq \gamma_{\Omega}(B)$).
The \textit{convex hull/closure operator} is defined on $\mathbb{R}^d$ with $d \in \mathbb{N}.$ This operator maps each subset $A \subseteq \mathbb{R}^d$ to the smallest convex set that contains $A$. Note that it defines indeed a closure operator on $\mathbb{R}^d$. 

\section{Definition of the union-free generic depth function on posets and some computational aspects}\label{app:def ufg depth}

In the following, we define the union-free generic (ufg) depth on the set of partial orders (posets), see appendix~\ref{app: prem_def} and \cite{pmlr-v215-blocher23a}. Therefore, let $M$ be a finite set and $\Pcal$ the set of all possible posets on $M$.

Depth functions can be regarded as a generalization of the univariate median and quantiles to multidimensional spaces. They provide a natural ordering of data points from center to outwards based on an observed data cloud or underlying distribution. There have been several different depth functions defined, like the simplicial depth on $\mathbb{R}^d$, see~\cite{liu90}, or depths on functional data, see~\cite{gijbels17}.  

The union-free generic (ufg) depth function developed by~\cite{pmlr-v215-blocher23a} is based on the simplicial depth on $\mathbb{R}^d$, see~\cite{liu90}. As described in section~\ref{sec:method}, the empirical simplicial depth uses the $d+1$ simplices with observed edges and computes the proportion of simplices that contain the point of interest. Thus, it is based on the $d+1$ simplices, where all edges are observed points $\{x_1, \ldots, x_{n} \} \in \mathbb{R}^d$ with $n \in \mathbb{N}$.  Here, \say{contain} means that the point of interest is inside the simplex. Hence, it lies in the output of the convex hull/closure operator,  see appendix~\ref{app: prem_def}. In particular, from the perspective of the convex hull operator, the edges of $d+1$ simplices are special subsets of $\mathbb{R}^d$. More precisely, these subsets are non-trivial in the sense that the output of the convex hull operator is a proper superset. They are minimal in that we cannot delete one point without changing the output. Moreover, they cannot be divided without loss of information. This means that dividing these edges into proper subsets and looking at the union of the output of the convex hull operator applied to these subsets yields a different set than applying it directly to the entire set, see Carath{\'{e}}odorys theorem on convex sets, see~\cite{eckhoff93}.
In~\cite{pmlr-v215-blocher23a} the authors describe this connection in more detail and exploit it in the definition of ufg depth.  

Since the ufg depth is an adaptation of the simplicial depth, \cite{pmlr-v215-blocher23a} starts by defining a closure operator, see appendix~\ref{app: prem_def}, on $\mathcal{P}$:
\begin{align*}
	\gamma\colon 
		2^{\Pcal} \to 2^{\Pcal}, \quad
		P \mapsto \left\{p \in \Pcal \mid \bigcap\limits_{
  \tilde{p}\in P}\tilde{p} \subseteq p \subseteq \bigcup\limits_{\tilde{p} \in P}\tilde{p} \right\}.
\end{align*}
 The next step is to adapt the $d+1$ simplices to this new framework. Thus, the non-triviality and the union-freeness are defined more general by~\cite{pmlr-v215-blocher23a}, and one obtains the following two conditions for $P\in\Pcal$: 
\begin{enumerate}
    \item[(C1)] $P \subsetneq \gamma(P),$
    \item[(C2)] there does not exist a family $(A_i)_{i \in \{1, \ldots, \ell\}}$ such that for all $i \in \{1, \ldots, \ell\}$ $A_i\subsetneq P$ and $\bigcup_{i \in \{1, \ldots, \ell\}} \gamma(A_i) = \gamma(P).$
\end{enumerate}
The authors showed that by applying these two conditions to $\mathbb{R}^d$ together with the convex hull operator, one ends up with the $d+1$ simplices. Thus, \cite{pmlr-v215-blocher23a} set $$\Sscr = \left\{P \subseteq\Pcal \mid \text{ Condition } (C1) \text{ and } (C2) \text{ hold} \right\}$$ 
and say that it consists of union-free generic sets.
\begin{example}\label{exampl: ufg_1}
    We consider a subset of the observed posets given by the benchmarking suite \texttt{DeepOBS} by~\cite{schneider2019deepobs}, see section~\ref{sec:results}. These are
    \begin{align*}
        p_1 &= \{(\text{SGD}, \text{Momentum})\}, \\
        p_2 &= \{(\text{SGD}, \text{Adam})\}, \: \text{ and}\\
        p_3 &= \{ (\text{Momentum},\text{SGD}), (\text{Momentum}, \text{Adam}) , (\text{SGD}, \text{Adam})\}.
    \end{align*}
    Note that $p_3$ is the poset on the left of the figure~\ref{fig:max_min_depth_poset:main}. Then, since $\gamma(\{p_1, p_2, p_3\}) =\{p_1, p_2, p_3\}$ is trivial and gives no more information, we get $\{p_1, p_2, p_3\} \not\in \Sscr$. (Note that it is not trivial when it contains a poset that is not observed.) As $p^*=\{(\text{SGD}, \text{Momentum}) , (\text{SGD}, \text{Adam})\} \in \gamma(\{p_1, p_2\}) $ and no proper subset of $\{p_1, p_2\}$ contains the poset $p^*$ in the output of its closure operator, we get that $\{p_2, p_3\} \in \Sscr$. 
\end{example}

Now, analogous to simplicial depth, the union-free generic depth function of a partial order $p$ is the weighted proportion of sets $S \in \Sscr$ containing $p$ in its closure. The weight comes from the positive probability of observing the same poset more than once
\begin{definition}
Let $p_1, \ldots, p_n \in \Pcal$ be a sample with corresponding empirical probability measure $\nu_n$ (equipped with the power set as $\sigma$-field). Then, the \textit{(empirical) union-free generic (ufg) depth} is given by
\begin{align*}
	D_n \colon \begin{array}{l}
		\Pcal \to [0,1] \\
		p \mapsto  \begin{cases}
            0, \quad &\text{if for all } S\in \Sscr\colon \prod_{\tilde{p} \in S}\nu_n(\tilde{p}) = 0\\
            c_n \sum_{S \in \Sscr, p \in \gamma(S)} \prod_{\tilde{p} \in S}\nu_n\left(\tilde{p}\right), &\text{else}
              \end{cases}
	\end{array}
\end{align*}
 with $c_n = \left(\sum_{S \in \Sscr} \prod_{\tilde{p} \in S}\nu_n\left(\tilde{p}\right)\right)^{-1}$, see~\cite{pmlr-v215-blocher23a}. 
\end{definition}
 Note that since $\nu_n(p) = 0$ if $p \in \Pcal$ is not observed, we can restrict the set $\Sscr$ to $\Sscr_{obs} = \{S \in \Sscr \mid S \subseteq \{p_1, \ldots, p_n\}\}$ consisting only of the observed posets. 
\begin{example}
    Recall example~\ref{exampl: ufg_1}, assume that only $p_1, p_2$ and $p_3$ were observed (without duplicates) and let $\nu_3$ be the corresponding empirical probability measure. By the remark after the definition of the ufg depth, we get that it is sufficient to consider only all union-free and generic sets that can be obtained by restricting to the observed posets. Thus, we obtain that $\Sscr_{obs} = \{\{p_1, p_2\}, \{p_2, p_3\}, \{p_1, p_3\}\}.$ With this, we get that the ufg depth value of all three observed posets is equal to $2/3$.
\end{example}
In~\cite{pmlr-v215-blocher23a}, the authors also introduce a population version. In addition, they showed some properties that the union-free generic depth function satisfies. This consists of properties such as how duplicates affect the result, what kind of posets have a depth of zero (e.g., when the poset has a preference structure that does not appear in any observed poset), and showing that the ufg depth approach cannot be reduced to a function based only on the pairwise comparisons. Furthermore, the authors prove that the ufg depth is consistent and therefore converges to the population version of the ufg depth under the independent and identically distributed (i.i.d.) assumption.

Finally, we want to discuss some computational aspects. The space of all possible observable posets grows with the number of elements $\# M = n$. Solving the exact number of posets for a set $M$ with $n = \# M$ is NP-hard, a lower bound is given by $2^{n^2/4}$, see~\cite{kleitman70}. Although the ufg depth function is defined over all possible posets, computing the depth of all possible posets can become computationally intensive. This is only needed if we are interested in inference, see appendix~\ref{app:benchmark-suite-design}. Another computationally intensive aspect is testing whether a subset satisfies conditions (C1) and (C2). In \cite{pmlr-v215-blocher23a} the authors explain in detail how this can be reduced. However, the larger $n = \# M$ is, the longer the computation takes. This can be seen in the runtime of our three examples: \texttt{DeepOBS} (approximate runtime: 1 second), Multi-Objective Evolutionary Algorithms (approximate runtime: 10 seconds), and BBOB (approximate runtime: 7 hours). Note that not only is the $\# M$ different, but also the number of test functions is larger in the BBOB suite.

\section{Results on BBOB Suite} \label{results-bbob}

Many researchers rely on the Black-box Optimization Benchmarking (BBOB) suite of test functions for comparing new optimizers against existing ones. We focus on results and evaluations from the BBOB 2009 and 2010 workshops on 24 noiseless functions in continuous domain in a black-box optimization scenario, see~\cite{hansen2010comparing}. The aim of the test function selection was to provide difficult and commonly faced optimization problems. These test functions are scalable with dimensions 2, 3, 5, 10, 20, and 40, see~\cite{mersmann15}. To evaluate the performance we use the calculations of~\cite{mersmann15}. We focus on comparing the 11 algorithms chosen in~\citet[Section 3.4]{mersmann15} as representatives of 11 algorithm groups. These are RANDOMSEARCH, POS, G3PCX, MOS, BIPOP-CMA-ES, IPOP-CMA-ES, FULLNEWUOA, (1+2\_m\^{}s) CMA-ES, iAMALGAM, BFGS and NELDER-DOERR. We also decided to limit our analysis to dimension 2. As performance measures, we use the expected runtime for precision level (distance to optimum) 0.001, see~\cite{auger05}, and the number of precision levels achieved overall.

Before discussing the ufg depth based on the posets given by test functions in the BBOB suite, we take a closer look at the construction of the partial orders. For each test function, we say that optimizer $i$ outperforms optimizer $j$ iff at least one criterion states that optimizer $i$ is better and all other criteria say that optimizer $i$ is not worse. If two criteria contradict each other in the sense that one criterion says optimizer $i$ is better and another criterion states the opposite, then these two optimizers are incomparable. More precisely, we say that based on this set of criteria, the optimizers cannot be compared. However, this construction does not cover the possibility that two optimizers are equal in all criteria. In this situation, the optimizers cannot be distinguished on the basis of the criteria used. So they are indifferent. Note that this is actually a distinction from incomparability, where the individual criteria do not agree on the performance order. One can say that in the case of indifference, optimizer $i$ is preferred over optimizer $j$, and vice versa, optimizer $j$ is preferred over optimizer $i$. This precisely defines only the weaker structure of a preorder, and not a partial order (antisymmetry is missing, see definitions in appendix~\ref{app: prem_def}). Thus the first naive approach, to include this indifference also as incomparability to the posets, does not do justice to the different interpretive backgrounds.\footnote{Generalizing the ufg depth to the set of preorders could be future work.}
\begin{figure}
\centering
\begin{subfigure}{.5\textwidth}
  \centering
  \includegraphics[width=.9\linewidth]{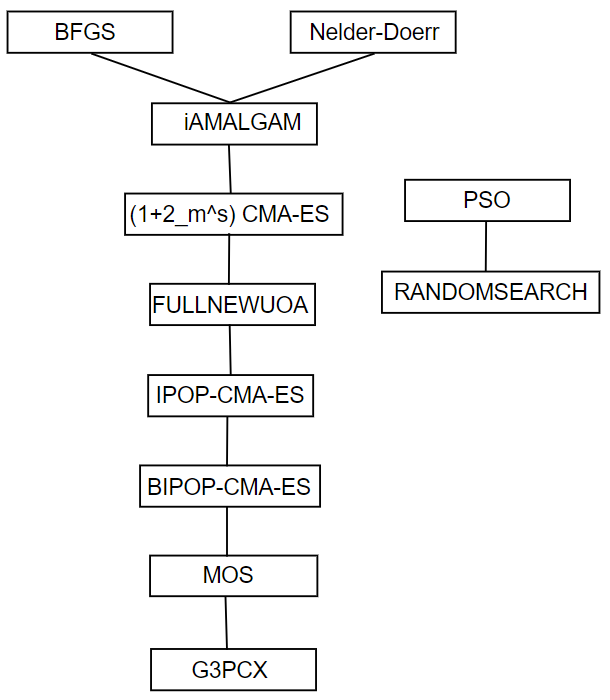}
  \label{fig:bbob_max_depth}
\end{subfigure}\hfill
\begin{subfigure}{.5\textwidth}
  \centering
  \includegraphics[width=.9\linewidth]{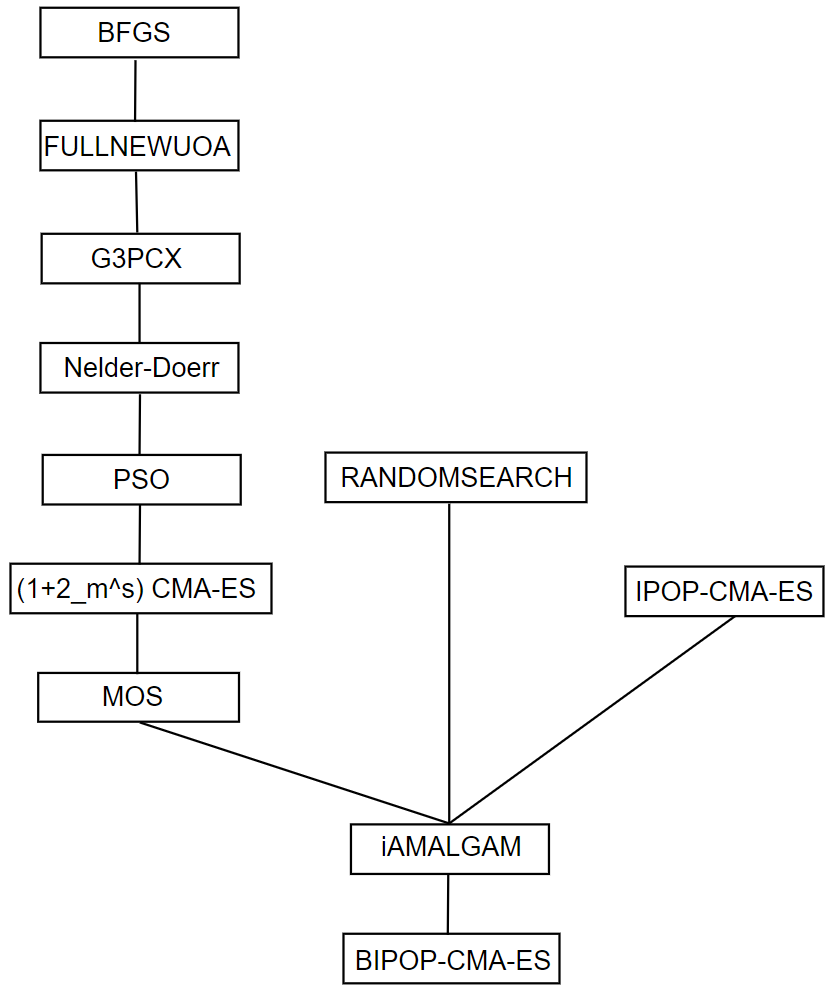}
  \label{fig:bbob_min_depth}
\end{subfigure}
\caption{BBOB suite based on dimension 2: Poset corresponding to the maximal (0.21, left) and minimal (0.11, right) ufg depth value.}
\label{fig:bbob}
\end{figure}

The problem of indifferent optimizers arises with the 11 optimizers given by~\cite{mersmann15} and the noiseless test functions in the BBOB suite with dimension 2. For test functions 1, 18, and 24, we have that at least two of the optimizers BFGS, FULLNEWUOA, RANDOMSEARCH, IPOP-CMA-ES, and G3PCX have equal criteria values. Therefore, we restrict our analysis of the ufg depth function to the posets given by the test functions without 1, 18, and 24.\footnote{If we accept the abuse of incomparability and add indifference as well as incomparability and run the analysis on all 24 test functions, we get a similar result.} Thus, we have 21 test functions resulting in 21 posets describing the performance order of the 11 optimizers where 20 are unique. Applying the ufg depth on these posets, we obtain a maximum depth value of 0.21 and a minimum depth value of 0.11. Since, in general, the ufg depth maps to $[0,1]$, this indicates that there is no poset that is strongly supported because it lies in many observed union-free generic sets. Thus, each poset contains a dominance or non-dominance structure that is not supported by observed poset subsets. This result may be due to the large variety of different optimization problems reflected by the test functions. Nevertheless, a discussion still can give some insight into the performance order structure of the optimizers and the corresponding test functions. 

The maximum ufg depth value is attained for the poset in figure~\ref{fig:bbob} (left). This is a duplicated one and the corresponding test functions are 10 and 11. Note that the outperformance of NELDER-DOERR and BFGS over all other optimizers except (PSO, RANDMOMSEARCH, and FULLNEWUOA) exists within the posets with the 8 highest ufg depth values. The observation of the dominance of NELDER-DOERR for the test functions restricted to 2 dimensions is consistent with~\cite{hansen2010comparing}. Besides the poset in figure~\ref{fig:bbob} (left) being the one where the dominance order (edges between the optimizers) is most supported, it is also the poset where the incomparability or non-dominance (non-edges) between two optimizers has the most typical order. Interestingly, we observe that the 9 most central posets all agree that RANDOMSEARCH is not dominated by any optimizer other than PSO. Note that this does not mean that RANDOMSEARCH dominates any other optimizer, only that for all corresponding test functions and for each optimizer there is at least one criterion where RANDOMSEARCH is better than the other optimizer. The poset in figure~\ref{fig:bbob} (right) has the minimum ufg depth value of 0.11. Thus, the corresponding test function 22 produces a performance order of the optimizer that is more outlying compared to all other performance orders produced. However, since all ufg depth values are quite low, it is questionable whether this can really be considered an outlier compared to all others, or whether all posets differ from each other to some extent.

\begin{figure}
    \centering
    \includegraphics[scale = 0.7]{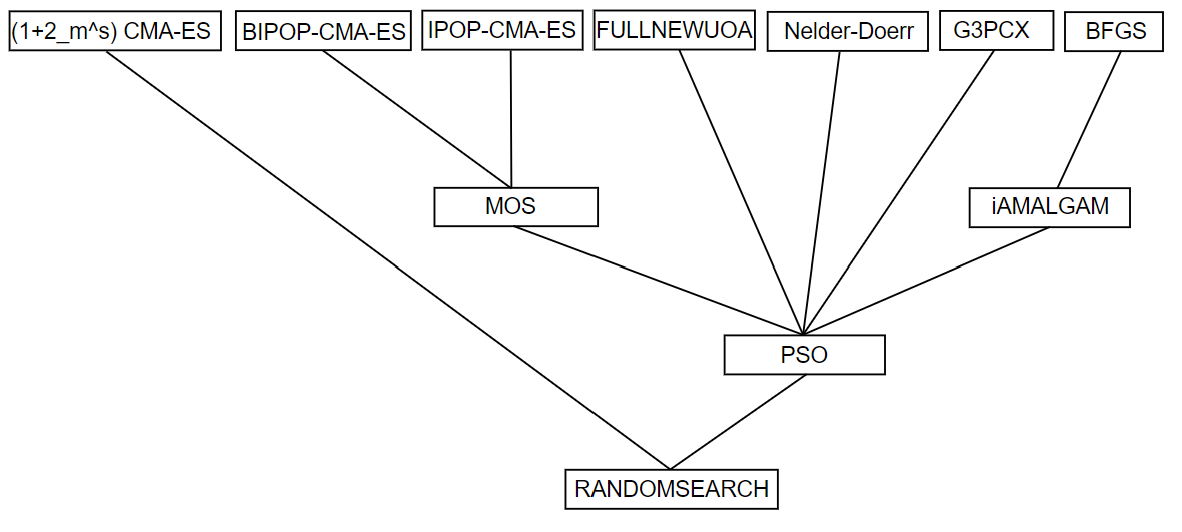}
    \caption{BBOB suite based on all dimensions: Three posets with the highest depth value have this dominance order in common, i.e., RANDOMSEARCH being dominated by all other optimizers is true for the three most central posets.}
    \label{fig:bbob_all_dim}
\end{figure}

Finally, we want to compare our analysis with \citet[p. 176]{mersmann15}. In that article, the authors analyzed the optimizers based on all six dimensions. Here, the aim was to obtain an overall total order of the optimizers. This was done with Borda consensus ranking, see appendix~\ref{app: related work}. Since the above analysis reflects the performance based only on dimension 2, we further evaluated the ufg depth function and posets, where for each of the test functions we use the expected running time of each dimension as a performance measure. We observe that the three posets corresponding to the three highest ufg depth values differ strongly and only agree on the dominance order in figure~\ref{fig:bbob_all_dim}. 
The main takeaway is that the three posets that are most central according to ufg depth only agree on random search being dominated by all other optimizers, while optimizer PSO is dominated by all expect random search. Optimizers MOS and iAMALGAM both dominate PSO and random search, but are incomparable w.r.t each other and are dominated by all remaining optimizers, which are in turn incomparable among each other.
 Since the dominance order falls apart quickly, it is questionable whether an overall order as suggested in~\citet[p. 176]{mersmann15} is reasonable.




\section{Results on Multi-Objective Evolutionary Algorithms}\label{sec: results-multi-objective-EA}

\begin{figure}[H]
  \begin{center}
    \includegraphics[width=0.8\textwidth]{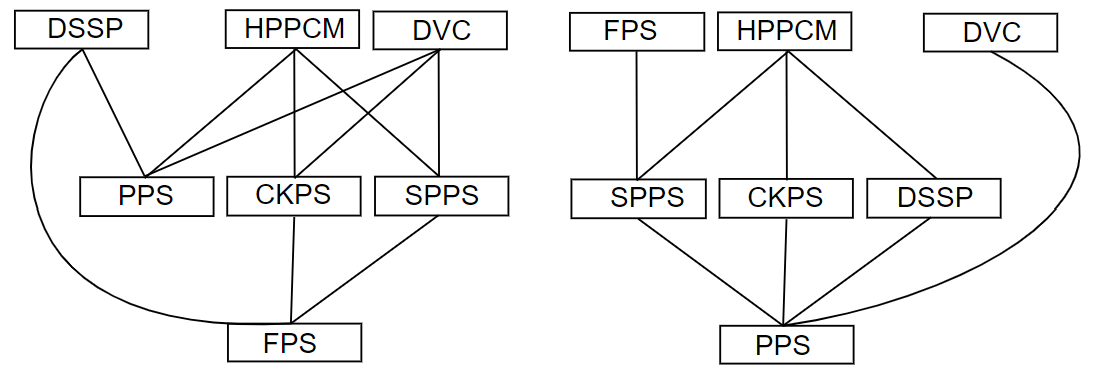}
    \caption{Multi-objective evolutionary algorithms: Orderings of optimizers corresponding to highest ($0.39$, left) and lowest ($0.17$, right) ufg depth.}
    \label{fig:max_min_depth_poset:evol_algorith}
  \end{center}
\end{figure}

We apply our framework on recently published benchmarking results for dynamic multi-objective evolutionary algorithms \citep{wu2023dynamic}. The proposed algorithm by \citet{wu2023dynamic} called DVC is compared against six state-of-the-art multi-objective evolutionary algorithms on 13 test functions with respect to the mean inverted generational distance (MIGD)\footnote{The MIGD is based on the minimum sum of distances between solutions belonging to the actual Pareto front and the solutions generated by the algorithm.} at four different phases (4 criteria).\footnote{Notably, \citet{wu2023dynamic} also benchmark with respect to the mean hypervolume difference at four different stages. For ease of demonstration, we abstain from including these additional criteria. Note that benchmarking according to 8 criteria would increase the number of incomparabilities.} This results in 13 (unique) posets describing the relation between the optimizers' 4 performance criteria on each of the 13 test functions. Figure~\ref{fig:max_min_depth_poset:evol_algorith} illustrates the posets with the highest and lowest depth value, respectively. 
The two depicted posets correspond to those 2 functions from the 13 considered test functions that entail the most typical relation of the optimizers' performances and the most outlying (atypical) one, respectively.

It becomes evident that the proposed evolutionary algorithm DVC is superior to all but two optimizers (namely, HPPCM and DSSP) on the most typical test function. On the most atypical one, it only dominates PPS and is incomparable to all others. Notably, this very simple illustration of our framework is enough to detect an oversimplified interpretation by \citet[page 10]{wu2023dynamic}: \say{From an overall perspective, the DVC algorithm performs well in all phases, although it is slightly inferior to HPPCM in the FDA3 problem and the dMOP3 problem (2 of the 13 test functions, the authors), but performs better in all other test problems}. Figure~\ref{fig:max_min_depth_poset:evol_algorith} (left graph) reveals that this statement is at least misleading. The DVC algorithm is in fact outperformed with respect to at least one criterion by DSSP. Otherwise, the two would not appear as incomparable in figure~\ref{fig:max_min_depth_poset:evol_algorith}. A brief glimpse at the tables reporting the criteria values for all test functions and all optimizers in \citet[tables 2-5]{wu2023dynamic} confirms this. In fact, it can be seen that DVC is outperformed by several more optimizers with respect to at least one criterion. 

\section{Outlook: Design and Curation of Benchmarking Suites} 
\label{app:benchmark-suite-design}

When designing and curating benchmarking suites, 
the question arises quite naturally as to what extent benchmarking results can be compared to each other. It might be of particular interest for benchmarking suite designers to gauge the diversity (dispersion) of the observed partial orderings.

We first note that the answer to this question depends on the definition of a benchmarking suite. While there is broad consensus that the latter entails the specification of test functions and evaluation criteria, it is not clear whether some tested optimizers are an integral part of a benchmarking suite, functioning as baselines for future benchmarking. The investigated \texttt{BBOB} benchmarking suite \citep{hansen2010comparing, hansen2021coco}, see section \ref{fig:bbob}, does not come -- to the best of our knowledge -- with a pre-specified set of baseline optimizers. The \texttt{deepOBS} benchmarking suite \citep{schneider2019deepobs}, however, comprises three deep learning optimizers as baselines (SGD, momentun, adam). 

If there are no specified baselines, our descriptive analysis of the benchmarking results says little if anything about the benchmarking suite \textit{as such}, because they are subject to the actually used optimizers in the benchmark. Moreover, the space of possible observable posets changes with the number of optimizers. More specifically, the cardinality of the possible observable posets grows with lower bound $2^{n^2/4}$ with $n$ the number of items/optimizers, see~\cite{kleitman70}.\footnote{Of course, the number of criteria used can restrict the number of possible observable posets as well, see order dimensions~\cite{yannakakis82}.} Moreover, the different cardinality of possibly observable posets strongly affects the \textit{a priori} probability of observing similar posets, e.g., when we have a discrete uniform distribution on posets with 3 optimizers versus on posets with 11 optimizers, it is more likely to observe 3 duplicated posets out of 8 observations for 3 optimizers than for 11 optimizers.
However, if the benchmarked optimizers are considered a property of the benchmarking suite, our descriptive analysis is more meaningful: The depth values describe all partial orderings produced by the benchmarking suite. Neither more test functions (\say{observations}) nor more optimizers (\say{features/items}) are observable for a so-defined benchmarking suite. Statistically speaking, this translates to a complete survey, where the sample equals the population. From the perspective of a benchmarking suite designer, the ufg depth function values (and their dispersion, measured e.g., by their range) are thus a sensible tool to assess the diversity of the partial orders produced by the suite. They might also be used to compare their diversity to the results of other benchmarking suites, keeping in mind that they refer to different populations.

However, a benchmarking practitioner typically uses the benchmarking suite to compare a newly proposed optimizer with existing ones. She is thus more interested in \textit{inferential} statements such as \say{Which test functions produce more likely a typical partial ordering of the old and new optimizers combined, and which produce an atypical order?} Practically speaking, she wants to get rid of test functions without changing the result. Since the new optimizer is unobserved with no prior knowledge, this question cannot be answered (by any ranking method). We can only restrict ourselves to some heuristics about how the distribution on the test functions changes for different optimizers. For example, we could assume that the test functions that produce typical or atypical poset structure on the benchmarking suite behave similar if we add a further optimizer from the same class of optimizers (e.g., evolutionary algorithms) to the benchmarking suite. This can make future experiments more time and energy efficient, while presumably not changing their results. 

A second question that concerns a benchmarking practitioner is: \say{On another test function, what is the most likely poset structure to be observed?} At this stage we cannot answer such questions because our analysis is descriptive only, see also section \ref{sec:results}. Future work will focus on statistical inference from posets. However, we want to point out that statistical inference will be difficult because the test functions, and thus the posets, can hardly be seen as independent and identically distributed over the space of all possible test functions/posets.





\section{Related Work}
\label{app: related work}

In this section, we provide some general background on benchmarking optimizers with respect to multiple criteria on a suite of test functions and review related work. Particular attention is paid to the notorious shortcomings of aggregating benchmarking results and how the presented approach avoids them. Before turning to concurring benchmark analysis methods in more detail, we start by a general description of the problem.

Comparing optimizers with respect to a single continuous criterion produces a complete ranking, i.e., a total order (see definition in appendix A). Aggregating several such orders arising from multiple criteria into one unique total order is a long-standing problem in social choice theory dating back to \cite{borda1781memoire} and \cite{marquis85}, still being subject of vivid discussion in economics. 
In the situation of this paper, the goal is to find a procedure/function that maps every possible set of total orders on any finite set $M$ to a total order. \say{Arrow's Impossibility Theorem}, see~\cite{arrow1950difficulty}, now states that this is not possible under natural desirable properties. These properties on the procedure, and especially on the resulting aggregated total order, contain the following three axioms among others.
For all three axioms, let $M \supsetneq \{a,b\}$ be a finite set with $\{p_1, \ldots, p_n\}$ ($n\ge 2$) being a set of total orders.

\begin{enumerate}
    \item If every ranking $p_i$ for $i \le n$ prefers $a$ over $b$, then the aggregated ranking shall prefer $a$ over $b$.
    \item 
    The aggregated preference of $a$ over $b$ shall not depend on changes of single total orders $p_i$ for $i \le n$ w.r.t. other items. 
    \item  The aggregation shall take into account all single rankings instead of following one predetermined ranking (\say{non-dictatorship}).
\end{enumerate}

Then \say{Arrow's impossibility theorem} states that for each procedure there exists a set of total rankings on $M$ such that the resulting aggregated total order does not fulfill all desired properties. For more details, see~\citep[chapter 7]{simon10}. 
Note that the second axiom implies the absence of cardinal information on the preferences. Cardinal information can be seen as some kind of strength between the comparisons, e.g., the difference of the estimated performance of one single measure between $a$ and $b$ is only 0.0001 instead of 1000. By allowing for such information (which is available in the typical case of continuous criteria) positive results for the aggregation problem are reachable, see \cite{bacharach1975group} for instance. However, one may also argue that due to incommensurability (e.g., different scalings, or variation of test functions' difficulties) of different continuous criteria such a cardinal information is not given. Using only ordinal information, there exist many methods/procedures to obtain one single aggregated total order, e.g., Borda ranking~\cite{dwork2001rank,borda1781memoire} or Kemeny-Snell method, see~\cite{kemeny62}. Further approaches can be found in~\cite{jansen18}. Note that these methods all have the problem of \say{Arrow's impossibility theorem}.

Going back to the benchmarking approach with multiple test functions and multiple performance criteria, the question of aggregating total orders arises twice. First, when for each individual test function we have a set of total orders (given by the individual criteria values), where the aggregation needs to represent the performance structure on that test function. Second, when an overall ranking of all performance orders of the test function is desired. Current approaches handle this in different ways. 

Despite the restriction of \say{Arrow's impossibility theorem}, established benchmarking schemes usually proceed by aiming at an aggregated overall total order, see \cite[section 5]{mersmann2010benchmarking} or \cite[section 2]{dewancker2016strategy}.
\cite{mersmann2010benchmarking,mersmann15} take into account the problem of the first aggregation on single test functions by integrating the observed rankings as partial orders. Nevertheless, the overall goal was to find one single total ranking describing the performance of the optimizers, which is achieved by consensus ranking. They critically discuss the downsides of such an aggregation, see \cite[section IV]{mersmann2010benchmarking} for instance: \say{The choice of consensus method is crucial and at the same time highly subjective. There is no right way to choose and therefore no one best algorithm. Our choice of algorithm will always depend on our choice of consensus method.} Similarly, the method of~\cite{dewancker2016strategy} results in an overall total order that does justice to the partial order character of each individual test function. There, the authors propose to derive final scores for each optimizer from the individual partial orders through a voting mechanism. In contrast, \cite{jansen2023statistical} returns an overall partial order. Here the authors work with so-called preference systems, which are an extension of the partial order to include the cardinal information, see also \cite{jsa2018,uaiall}. Note that \cite{jansen2023statistical} actually present a method to compare classifiers with respect to different quality criteria rather than optimizers, but their approach is principled and can be easily applied to optimizers.

Like \cite{dewancker2016strategy}, we consider partially ordered optimizers arising from multiple criteria for a single test function. The main difference to all the approaches above is that our goal is not to obtain a single partial order that is an aggregation, but rather to describe the entire distribution of possible partial orders representing the performance of the optimizers. This is done by giving each partial order a measure of how central and outlying it is. Most importantly, we can also observe structures that are outlying/atypical and therefore find test functions that produce atypical performance structures. In this way, we take into account the fact that, from our point of view, a single performance structure is not sufficient to describe the overall performance. Nevertheless, the partial order with the highest depth value is the most central partial order and can be seen as the most typical performance structure. With this, one can still compare our result with other benchmarking approaches. Appendix~\ref{results-bbob} gives an example of the method comparison.

\end{document}